\documentclass[letterpaper, 10 pt, journal, twoside]{IEEEtran}

\IEEEoverridecommandlockouts                              

\usepackage{microtype}
\usepackage{subfigure}

\usepackage{mathtools}
\usepackage{amsthm}

\theoremstyle{plain}

\theoremstyle{definition}

\theoremstyle{remark}

\usepackage{graphicx}
\usepackage{amsmath}
\usepackage{amssymb}
\usepackage{booktabs}
\usepackage{comment}

\usepackage{colortbl}
\usepackage{color,soul}
\usepackage{adjustbox}
\usepackage{xcolor}
\usepackage{bbding}
\usepackage{xspace}
\definecolor{lower_metric}{rgb}{0.839,0.651,0.596}
\definecolor{higher_metric}{rgb}{0.65,0.765,0.914}
\definecolor{StandardYellow}{rgb}{1,1,0.4}
\definecolor{LightYellow}{rgb}{1,1,0.88}
\definecolor{StandardCyan}{rgb}{0.4,1,1}
\definecolor{LightCyan}{rgb}{0.88,1,1}

\usepackage{physics}

\usepackage{pifont}
\newcommand{\cmark}{\ding{51}}
\newcommand{\xmark}{\ding{55}}

\usepackage{multirow}

\usepackage{enumitem}

\definecolor{mygreen}{rgb}{0.13,0.545,0.13}

\usepackage[linesnumbered,ruled,vlined,noend]{algorithm2e}

\usepackage{arydshln}

\usepackage[pagebackref,breaklinks,hidelinks,colorlinks]{hyperref}
\hypersetup{
  colorlinks,
  citecolor=blue,
  linkcolor=red,
  urlcolor=violet}
\usepackage[capitalize,noabbrev]{cleveref}

\crefname{section}{Sec.}{Secs.}
\Crefname{section}{Section}{Sections}
\Crefname{table}{Table}{Tables}
\crefname{table}{Tab.}{Tabs.}

\title{\LARGE \bf
SemHint-MD: Learning from Noisy Semantic Labels \\for Self-Supervised Monocular Depth Estimation
}

\author{Shan Lin$^{1}$, Yuheng Zhi$^{1}$, and Michael C. Yip$^{1}$
\thanks{$^{1}$S. Lin, Y. Zhi, and M.C. Yip are with the Department of Electrical and Computer Engineering, University of California San Diego, La Jolla, CA 92093, USA
        {\tt\small \{shl102, yzhi, yip\}@ucsd.edu}}%
}

\begin{document}

\maketitle
\thispagestyle{empty}
\pagestyle{empty}

\begin{abstract}
Without ground truth supervision, self-supervised depth estimation can be trapped in a local minimum due to the gradient-locality issue of the photometric loss. 
In this paper, we present a framework to enhance depth by leveraging semantic segmentation to guide the network to jump out of the local minimum. 
Prior works have proposed to share encoders between these two tasks or explicitly align them based on priors like the consistency between edges in the depth and segmentation maps. 
Yet, these methods usually require ground truth or high-quality pseudo labels, which may not be easily accessible in real-world applications. 
In contrast, we investigate self-supervised depth estimation along with a segmentation branch that is supervised with noisy labels provided by models pre-trained with limited data. 
We extend parameter sharing from the encoder to the decoder and study the influence of different numbers of shared decoder parameters on model performance. 
Also, we propose to use cross-task information to refine current depth and segmentation predictions to generate pseudo-depth and semantic labels for training. 
The advantages of the proposed method are demonstrated through extensive experiments on 
the KITTI benchmark 
and a downstream task for endoscopic tissue deformation tracking. \footnote{We will release our implementation upon publication.}

\end{abstract}

\section{Introduction}

\IEEEPARstart{G}{eometrical} and semantic scene understanding have been critical needs in many robotic or computer vision applications \cite{chen2019suma++, doherty2020probabilistic, lin2022semantic, fan2022blitz}. The depth and semantic maps are two of the most commonly used modalities.
In these scenarios, depth is often acquired using expensive devices like LiDAR and the semantic maps are assumed to be near-ideal. Such setups can be either cost-prohibitive or impractical in real-world applications. 
For example, conventional depth sensors are generally unavailable for capturing surgical scenes, since endoscopic cameras are incredibly small and prohibit space for structured light or time-of-flight hardware \cite{khan2020deep, huang2022self}. 
Also, the generalizability of current segmentation models is still limited, so recent works have moved to more realistic scenarios where only noisy pseudo-labels are used for training, to reduce the need for manual data labeling \cite{hicks2021endotect, wang2022semi}. 

\begin{figure}[!t]
\vspace{0.5em}
\centering
\includegraphics[width=0.99\linewidth]{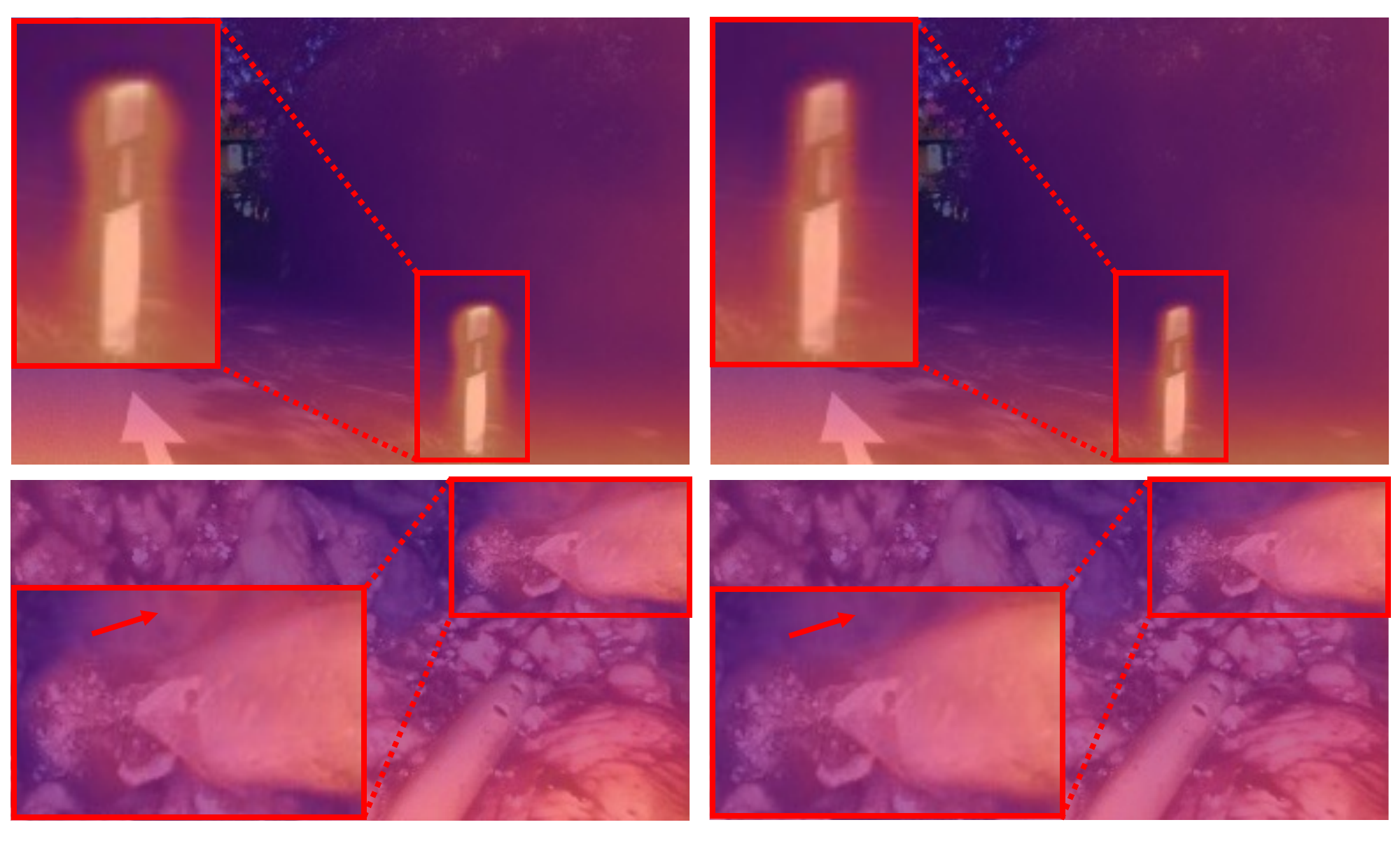}
\vspace{-2em}
\caption{The proposed method can improve depth near object boundaries. Left: Our baseline \cite{watson2019self}. Right: SemHint-MD (ours). The top-left image shows a typical ``bleeding artifact" (see Section \ref{sec:related_work} for details), while our method can alleviate this artifact, as shown in the top-right image.} \label{fig:front}
\vspace{-1em}
\end{figure}

Moreover, depth and segmentation are not only both highly needed in downstream tasks, but they also share many characteristics – they both are dense prediction tasks, are supposed to produce clear object boundaries, etc. And more importantly, depth and segmentation can complement each other. For example, current segmentation usually offers better object boundaries than self-supervised depth, which can provide hints for depth estimation, while depth maps provide 3D information like spatial relationships between objects, which can potentially help segmentation. 
There are a stream of methods look into multi-task learning (MTL) for depth and segmentation, which can be roughly categorized as implicit and explicit methods. 
Implicit methods share model parameters between the two tasks, allowing the model to learn from different aspects of the data by receiving supervision from different tasks \cite{kendall2018multi, jiao2018look, chen2019towards, zhang2019pattern}. 
Recent works \cite{zhu2020edge, wang2020sdc} suggest that implicit methods have not fully utilized the connections between depth and segmentation, such as the object boundaries shared between them. 
Thus, they explicitly encourage larger depth gradients to align with borders in the semantic map \cite{zhu2020edge, wang2020sdc}, or encourage disparity smoothness within each semantic region \cite{zama2018geometry, chen2019towards}. 
Many of these MTL methods require some, if not all, tasks come with ground truth or high-quality pseudo labels.

In contrast, we develop our MTL approach under a more realistic configuration, where depth estimation is self-supervised and the segmentation task only has access to noisy pseudo labels. 
This approach uses semantic information to mitigate the negative impact brought into the learning process by the limitation of current self-supervised losses for depth, so we call the proposed method as {\bf SemHint-MD}, which stands for {\bf Sem}antic {\bf Hint} for self-supervised {\bf M}onocular {\bf D}epth estimation. We first evaluated SemHint-MD on the widely tested KITTI benchmark. Additionally, one area where this setup is particularly relevant is endoscopic surgery, where accurately reconstructing 3D geometry and inferring anatomy labels in surgical videos can
guide surgeons in 
diagnosis and treatment. 
To this end, we also demonstrate this method on 
a downstream surgical scene reconstruction problem. 
Our main contributions are as follows: 
\begin{itemize}[leftmargin=1em]
\item We extend the widely adopted encoder sharing strategy to also sharing decoder layers and investigate its influence on model performance. 

\item We propose a mutual refining algorithm that incorporates complementary information from depth and segmentation to mitigate the limitation of self-supervised losses for depth and the negative impact of noisy semantic labels, leading to better depth estimation.

\item We demonstrate that SemHint-MD can provide shorter inference delay to a downstream surgical scene reconstruction task and allow it to achieve better performance. 
\end{itemize}

\section{Related Work} \label{sec:related_work}

\subsection{Self-supervised Depth Estimation} 
The high cost of ground truth collection led to many self-supervised depth estimation models \cite{godard2019digging, watson2019self, yin2018geonet}. Such models are usually trained by minimizing the photometric loss between real images and images that are warpped from different viewpoints or times based on the predicted depth maps. 
Yet, photometric loss suffers from the well-known gradient-locality issue, so the network can be stuck in a local minimum. 
Also, the photometric loss will provide erroneous supervision to pixels that do not have correspondences between images captured from different viewpoints or at different times (due to, e.g., occlusions). 
In stereo pairs, such occlusions usually appear near object boundaries, where some background points are occluded by the foreground points and the self-supervised photometric loss will force the model to find a non-occluded, similar background point near the object boundary, which causes the depth of the foreground object expands into the background region (named``bleeding artifact") \cite{zhu2020edge}. Visually, in the depth map, there will be a foreground object region that is larger than its true shape, the top-left image in Fig. \ref{fig:front} shows a typical example of ``bleeding artifact".
In this paper, to overcome these limitations, we extract approximate depth values of the occluded regions or regions with unreliable depth from their nearby high-confident depth values of the same semantic class to serve as pseudo-ground truth, guiding the network to converge to better depth. 

\subsection{Multi-task Learning for Depth and Segmentation} 
MTL between depth and tasks that are mathematically related to depth, such as normal estimation \cite{yang2018unsupervised, li2021structdepth}, 
has been investigated to propagate supervision and regularization between tasks by converting predictions across them. 
Whereas, equations that can infer segmentation map from depth map, or vice versa, are not available. 
Thus, many works barely share encoder weights between depth and segmentation \cite{mousavian2016joint, kendall2018multi, jiao2018look, chen2019towards, zhang2019pattern}. 
To better leverage cross-task information, recent works proposed to explicitly encourage object boundary consistency between depth and segmentation predictions \cite{zhu2020edge} and disparity smoothness within each semantic region \cite{zama2018geometry, chen2019towards}. 
The semantic information was also utilized in a way similar to photometric loss, where the segmentation map of a source image is warped to the target view, and the network is trained by encouraging its consistency with the segmentation map of the target image \cite{chen2019towards}. 
The inconsistency between the warped and real segmentation maps was also used to identify dynamic objects and excluded them when calculating the photometric loss \cite{klingner2020self}. 
However, these approaches usually require ground truth or high-quality pseudo labels, so that one task can be accurate enough for improving the other task. 
In contrast, we propose a mutual refining algorithm that takes noisy depth and segmentation maps (predicted depth and segmentation, and pseudo-semantic labels), and lets them refine each other to provide pseudo-depth and semantic labels to supervise the network. 

\begin{figure*}[t]
\centering
\includegraphics[width=0.9\linewidth, trim={0 0 0 0}, clip]{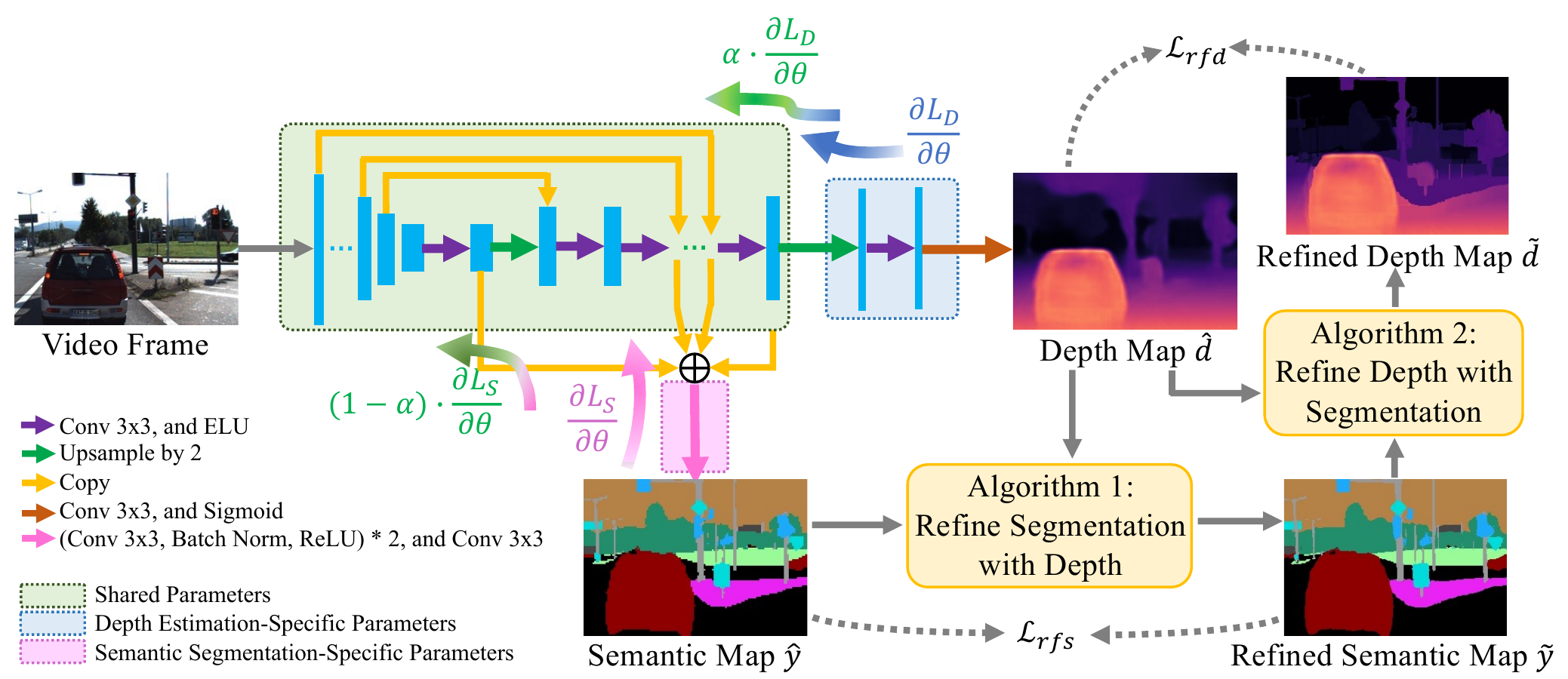}
\vspace{-1em}
\caption{Overview of the proposed framework. To keep this figure concise, only the depth hints loss is shown. Other commonly used self-supervised depth losses like photometric losses are not shown in this figure, see Section \ref{sec:related_work} for their details. } \label{fig:overview}
\vspace{-1em}
\end{figure*}

\subsection{Cross-modality Parameter Sharing}
Sharing parameters between different tasks in MTL can be roughly categorized into hard and soft parameter sharing \cite{ruder2017overview}. 
Hard parameter sharing usually has one shared encoder and individual decoder for each task \cite{ruder2017overview, zama2018geometry, zhang2021survey}. 
Few works extended hard sharing to the decoder and they all require ground truth labels. 
Hickson \emph{et al.} \cite{hickson2022sharing} showed that partially sharing the decoder can improve MTL performance with less computation. 
Chen \emph{et al.} \cite{chen2019towards} proposed a fully-shared model with task-specific embedding, which avoided the efforts of tuning the number of sharing layers. But this model requires to be run twice to get both depth and segmentation results. 
Soft parameter sharing, on the other hand, lets each task have its own parameters and enhance task-specific features by combining them with features of other tasks through a learnable linear combination \cite{misra2016cross, ruder122017learning, jiao2018look}.
Some recent works further adopted both hard and soft parameter sharing in one model \cite{jiao2018look, zhang2021survey}. 
In this paper, we focus on hard parameter sharing and investigate the influence of extending sharing from encoder to decoder, but the proposed methods can be extended to soft or hybrid parameter sharing models. 

\section{Method}

Fig. \ref{fig:overview} provides the overview of the proposed MTL model SemHint-MD. Stereo image pairs are used for training, and at test time, only monocular images are needed for prediction. 
We adapt a popular depth estimation network structure, Monodepth2 \cite{godard2019digging}, to include a new segmentation branch that shares the encoder and most decoder layers with the depth branch, whose architecture is shown in Fig. \ref{fig:overview} and illustrated in Section \ref{sec:structure}. 
When training the MTL model, we scale the gradients of the shared parameters to balance parameter update, which is explained in Section \ref{sec:training}. 
To mitigate the limitations of the photometric loss and the effect from noise in pseudo-semantic labels, 
we propose to refine the current predicted depth and segmentation maps by extracting cross-modality information from each other, which is introduced in Section \ref{sec:refine}, and further use the refined maps to supervise the training. 
All involved loss functions are introduced in Section \ref{sec:loss}. 

\subsection{Multi-Task Architecture} \label{sec:structure}

As discussed in Section \ref{sec:related_work}, most existing MTL models consist of a shared encoder and a separate decoder for each task. To further reduce the number of parameters, we explore the possibility of further sharing the decoder layers between depth estimation and semantic segmentation. 
As shown in Fig. \ref{fig:overview}, the depth estimation branch of the proposed MTL architecture is the same as Monodepth2 \cite{godard2019digging}, where ResNet \cite{he2016deep} is used as the encoder, and the decoder consists of several convolution layers that are followed by nearest upsampling layers, skip connections from the encoder, and ELU activation functions \cite{clevert2015fast}. 
The output of the depth branch is a disparity map $\sigma$ and it will be converted to a depth map $d$ by $d=1/(c_1\sigma+c_2)$ \cite{godard2019digging}. 
The segmentation branch only requires three extra convolution layers. Specifically, we concatenate the features from four decoder layers that are shared with depth estimation to aggregate information from different scales and then pass them to the three new convolutions for prediction. 
The first two convolution layers are both followed by a BatchNorm layer and ReLU activation, while the last convolution layer outputs the segmentation map normalized by the softmax function. 

\subsection{Gradient Scaling for Multi-Task Training} \label{sec:training}
A common approach to train a MTL model is to control the influence of different tasks by weighting their losses accordingly, but in this way, the gradients inside the task-specific parameters will also be scaled, while the shared parameters are trained with larger gradients that come from all tasks. 
To balance the training of task-specific and shared parameters, we follow \cite{garg2016unsupervised, klingner2020self} to only scale the gradients when they reach the shared parameters. Specifically, inside the shared parameters, the gradients that come from losses for depth estimation are scaled by a hyperparameter $\alpha$ and the gradients that come from losses for semantic segmentation are scaled by $1-\alpha$, as shown in Fig. \ref{fig:overview}.
The gradient scaling is implemented using the method proposed in \cite{ganin2015unsupervised}.

\subsection{Refine Segmentation and Depth} \label{sec:refine}

To reduce the effect of wrong labels in the pseudo-semantic ground truth on the training of the segmentation branch, we propose to merge the predicted segmentation map with the pseudo-ground truth map with the assistance of the depth information to generate a refined segmentation map. 
On the other hand, as introduced in Section \ref{sec:related_work}, photometric loss sufers from gradient locality and provides erroneous supervision in occluded regions. Thus, we use the refined segmentation map to help generate a refined depth map that can provide depth hints in regions with unreliable depth predictions. 
The refined depth and segmentation maps are then used to supervise training and the corresponding loss terms are introduced in Section \ref{sec:loss}. 

\subsubsection{Refine Segmentation with Depth} This refinement method is based on the intuition that the depth values of points on the same object are closer to each other than the depth values of points on different objects. 
The detailed procedure of this method can be found in Algorithm \ref{alg:depth_help_seg}. 
First, the predicted segmentation map $\hat{y}$ is compared with the pseudo-ground truth $y$ to split elements in $y$ to two sets: a high confident set $\mathbb{C}$ of pixels $c_i$ whose labels are the same as their corresponding pixels $\hat{y}_i$ in the predicted map $\hat{y}$, and a less reliable set $\mathbb{U}$ of pixels $u_i$ whose labels are different from their corresponding pixels $\hat{y}_i$ in $\hat{y}$ (Line 1-2). 
Then the following steps are executed iteratively: 1) seek a more reliable label $c^* \in \mathbb{C}$ for each $u_i \in \mathbb{U}$ from its neighborhood $\mathcal{N}({u_i})$, where the depth at $c^*$ is closest to the depth $d_i$ at $u_i$ (Line 6-7); 2) update $u_i$ if the difference between the depth of $u_i$ and $c^*$, $\bigtriangleup d^*$, is less than a threshold $\bigtriangleup d_{th}$, which indicates that $u_i$ and $c^*$ are more likely to belong to the same object (Line 8-9); and 3) move updated elements from $\mathbb{U}$ to $\mathbb{C}$ (Line 10-11). Finally, all elements are moved to $\mathbb{C}$ to generate $\tilde{y}$ (Line 12). 
The operations on $u_i$ are parallelized through 2D convolution and max pooling in PyTorch. 

\begin{algorithm}
\caption{Refine Segmentation with Depth}\label{alg:depth_help_seg}
\KwIn{Pseudo-semantic ground truth $y$; 
Predicted segmentation map $\hat{y}$; 
Predicted depth map $\hat{d}$; 
Threshold of depth difference $\bigtriangleup d_{th}$.}
\KwOut{Refined segmentation map $\tilde{y}$.}
$\mathbb{C}\gets\left\{c_i \in y|c_i=\hat{y}_i\right\}$\; $\mathbb{U}\gets\left\{u_i \in y|u_i\neq\hat{y}_i\right\}$\;
\While{$\mathbb{U}\neq\varnothing$}{
    \ForAll{$u_i \in \mathbb{U}$}{
        \If{$\exists c_j\in\mathbb{C}: c_j\in\mathcal{N}(u_i)$}{
            $c^* \gets \underset{c_j\in \mathcal{N}(u_i)}{\arg\min} |d_i-d_j|$\;
            $\bigtriangleup d^* \gets \underset{c_j\in \mathcal{N}(u_i)}{\min} |d_i-d_j|$\;
            \If{$\bigtriangleup d^*<\bigtriangleup d_{th}$}{
                $u_i \gets c^*$\;
            }
            $\mathbb{U} \gets \mathbb{U}\setminus\{u_i\}$\;
            $\mathbb{C} \gets \mathbb{C}\cup\{u_i\}$\;
        }
    }
}
Integrate all elements in $\mathbb{C}$ to generate $\tilde{y}$\;
\end{algorithm}


\begin{algorithm}[tb!]
\caption{Refine Depth with Segmentation}\label{alg:seg_help_depth}
\KwIn{Predicted depth map $\hat{d}$; 
Refined segmentation map $\tilde{y}$ from Algorithm \ref{alg:depth_help_seg};
Semantic classes $\mathbb{Z}$;
Target and source image pair ($I^t$, $I^s$); 
The pose of the left view with respect to the right view $T^{t\rightarrow s}$; 
Camera intrinsic matrices $K$.}
\KwOut{Refined segmentation map $\tilde{d}$.} 
$I^{s\rightarrow t} \gets {\bf B} (I^s,\boldsymbol\pi(\hat{d},T^{t\rightarrow s},K)) $ \hfill {$\triangleright$ {\bf B}: Bilinear sampling}\;
$y^t \gets {\bf S}(I^t)$ \hfill {$\triangleright$ {\bf S}: Segmentation branch}\;
$y^{s\rightarrow t} \gets {\bf S}(I^{s\rightarrow t})$\;
\ForAll{$z_k \in \mathbb{Z}$}{
    $\mathbb{C}^k\gets\left\{c_i^k \in \hat{d} |y^t_i=y^{s\rightarrow t}_i, \tilde{y_i}=z_k\right\}$\;
    $\mathbb{U}^k\gets\left\{u_i^k \in \hat{d}|y^t_i\neq y^{s\rightarrow t}_i,\tilde{y}_i=z_k\right\}$\; 
    \While{$\mathbb{U}^k\neq\varnothing$}{
        \ForAll{$u^k_i \in \mathbb{U}^k$}{
            \If{$\exists c^k_j\in\mathbb{C}^k: c^k_j\in\mathcal{N}(u^k_i)$}{
                $c^{min}_i \gets \underset{c^k_i \in \mathcal{N}(u^k_i)}{\min} c^k_i$\;
                $c^{max}_i \gets \underset{c^k_i \in \mathcal{N}(u^k_i)}{\max} c^k_i$\;
                $u^k_i \gets \max(\min(u^k_i, c^{max}_i), c^{min}_i)$\;
                $\mathbb{U}^k \gets \mathbb{U}^k\setminus\{u^k_i\}$\;
                $\mathbb{C}^k \gets \mathbb{C}^k\cup\{u^k_i\}$\;
            }
        } 
    }
}
Integrate all elements in $\bigcup\limits_k\mathbb{C}^k$ to generate $\tilde{d}$\;
\end{algorithm}

\subsubsection{Refine Depth with Segmentation} 
This depth refinement method is based on the insight that depth in challenging regions, such as object boundaries where occlusions usually happen, is close to depth near the object center, where the predictions are usually more reliable as the photometric loss can provide better supervision. 
The detailed procedure of this method can be found in Algorithm \ref{alg:seg_help_depth}. 
First, the source image $I^s$ is warped to the target view to generate $I^{s\rightarrow t}$ (Line 1). 
Since stereo image pairs are used for training, the model randomly selects one image from each image pair as the target image $I^t$ and predicts its depth, which will then be used to warp another image in the pair, \emph{i.e.}, the source image $I^s$, to the target view. 
Then the segmentation branch {\bf S} is used to predict segmentation $y^t$ of the target image $I^t$ and the segmentation $y^{s\rightarrow t}$ of $I^{s\rightarrow t}$ (Line 2-3). 
Ideally, correct depth and segmentation will lead to consistency between $y^t$ and $y^{s\rightarrow t}$ in non-occluded regions and inconsistency in occluded regions, because correct depth and segmentation will find the foreground point that covers the corresponding background point and lead to semantic inconsistency between the source and target view. 
As true-semantic ground truth is not available, we use the current segmentation branch to predict $y^t$ and $y^{s\rightarrow t}$ for this consistency checking. In this way, even if the segmentation predictions are incorrect, similar points are more likely to be given the same labels,  which achieves similar goals. 
Next, the elements $\hat{d}_i\in\hat{d}$ in the semantic consistent regions are sent to the high confident set $\mathbb{C}$, while the elements in the inconsistent regions are send to the less reliable set $\mathbb{U}$, and we further split $\mathbb{C}$ and $\mathbb{U}$ to subsets $\mathbb{C}^k$ and $\mathbb{U}^k$ to let each subset only consists of elements from the same semantic class $z_k$ (Line 5-6). Note that we use the refined segmentation map $\tilde{y}$ to split $\mathbb{U}$ and $\mathbb{C}$, as it is considered to be more reliable than both the pseudo-ground truth and current segmentation prediction due to the depth-aware refinement. 
For each class $z_k$, the following steps are executed iteratively: 
1) for each $u^k_i\in\mathbb{U}^k$ that has $c^k_j\in\mathbb{C}^k$ in its neighborhood $\mathcal{N}(u^k_i)$,  
find the minimum and maximum high confident depth $c^{min}_i$ and $c^{max}_i$ from $\mathcal{N}(u^k_i)$ (Line 10-11); 
2) clip 
$u^k_i$ to [$c^{min}_i$, $c^{max}_i$] (Line 12); and 
3) move $u^k_i$ from $\mathbb{U}^k$ to $\mathbb{C}^k$ (Line 13-14). 
Finally, all elements are moved to their corresponding high confident sets and the union of all $\mathbb{C}^k$ generates the output $\tilde{d}$ (Line 15).
Similarly, the operations are parallelized through 2D convolution and max pooling layers. 

It should be noted that the depth hints provided by this refinement method are still noisy, but we believe that because most of them are closer to the true depth than current predictions, these depth hints can guide the network to jump out of the local minimum and converge to a better minimum. 

\subsection{Loss Functions} \label{sec:loss}

The total loss consists of loss for the depth branch $L_D$ and loss for the segmentation branch $L_S$
\begin{equation}
L = \beta_1L_D + \beta_2L_S
\end{equation}
where $\beta_1$ and $\beta_2$ are the weights of the losses. 

\subsubsection{Losses for Depth Estimation}
The depth loss is comprised of four terms:
\begin{equation}
L_D=\lambda_{pe}L_{pe}+\lambda_{h}L_{h}+\lambda_{rfd}L_{rfd}+\lambda_{s}L_{s}
\end{equation}
where $L_{pe}$ is the photometric loss, $L_{h}$ is the classical depth hint supervision loss, $L_{rfd}$ is the refined depth supervision loss, $L_{s}$ is the smoothness loss, and $\lambda_{pe}$, $\lambda_{h}$, $\lambda_{rfd}$, and $\lambda_{s}$ are the weights of the losses. 

{\bf Photometric Loss} Following \cite{godard2017unsupervised, godard2019digging, watson2019self}, we use a combination of $L_1$ and
Structural Similarity Index (SSIM) to measure photometric loss: 
\begin{equation}
L_{pe} = \frac{\gamma}{2}(1-SSIM(I^t, I^{s\rightarrow t})) + (1-\gamma)\|I^t - I^{s\rightarrow t}\|_1
\end{equation}
where $\gamma=0.85$, ($I^t$, $I^s$) is a pair of target and source image. 
In this work, we randomly select one image in a stereo pair as $I^t$ and let another image as $I^s$. $I^{s\rightarrow t}$ is $I^s$ warped to the target view using the predicted depth of $I^t$. 
We adopt the multi-scale strategy \cite{godard2019digging} to predict depth maps at four resolution scales and average photometric losses over these scales. 

{\bf Classical Depth Hint Supervision Loss} 
Watson \emph{et al.} \cite{watson2019self} proposed to use an off-the-shelf stereo matching algorithm Semi-Global Matching (SGM) \cite{hirschmuller2007stereo} to generate depth maps $d'$ as labels for supervision, and the loss is calculated as 
\begin{equation}
    L_h = \log(1+|\hat{d}-d'|)
\end{equation}

{\bf Refined Depth Supervision Loss} 
This loss uses the proposed refined depth $\tilde{d}$ generated by Algorithm \ref{alg:seg_help_depth} to provide another supervision: 
\begin{equation}
    L_{rfd} = \log(1+|\hat{d}-\tilde{d}|)
\end{equation}

 {\bf Smoothness Loss.} 
 To reduce texture-copy artifacts in textureless regions, such as the tissue surface, we adopt the edge-aware smoothness loss $L_s$ \cite{godard2017unsupervised} to regularize the depth estimation to be locally smooth. 
 Different from other depth losses, the smoothness loss is calculated using the disparity maps $\sigma$ (\emph{i.e.}, inverse depth):
\begin{equation}
L_s = |\partial_x \hat{\sigma}^*|e^{-|\partial_x I|} + |\partial_y \hat{\sigma}^*|e^{-|\partial_y I|}
\end{equation}
where $\hat{\sigma}^*$ is the mean-normalized disparity.

\subsubsection{Losses for Semantic Segmentation}
The segmentation loss is comprised of two terms:
\begin{equation}
L_S=\lambda_{ps}L_{ps} + \lambda_{rfs}L_{rfs}
\end{equation}
where $L_{ps}$ is the cross-entropy loss \cite{taha2015metrics} between prediction $\hat{y}$ and pseudo-semantic ground truth $y$: 
\begin{equation}
L_{ps}(y,\hat{y}) = -\frac{1}{N}\sum_iy_i\log(\hat{y}_i)
\end{equation}
where $N$ is the number of elements in the segmentation map, $y_i$ is the $i$-th element of $y$. 
We use off-the-shelf models to infer segmentation maps for all training data to serve as pseudo-ground truth $y$.
$L_{rfs}$ is the cross-entropy loss between $\hat{y}$ and the refined depth map $\tilde{y}$ generated by Algorithm \ref{alg:depth_help_seg}:
\begin{equation}
L_{ps}(\tilde{y},\hat{y}) = -\frac{1}{N}\sum_i\tilde{y}_i\log(\hat{y}_i)
\end{equation}
$\lambda_{ps}$ and $\lambda_{rfs}$ are the weights to control the losses. 

\begin{table*}[t]
\caption{Performance Comparisons on KITTI Stereo 2015 Dataset \cite{geiger2012we} Eigen Splits \cite{eigen2014depth} Capped at 80 Meters.} \label{tab:kitti}
\vspace{-1em}
\centering
\begin{adjustbox}{width=\textwidth}
\begin{tabular}{l|c|cc|ccccccc}
\hline
Method & PP & Data & Resolution & \multicolumn{1}{>{\columncolor{lower_metric}}l}{Abs Rel$\downarrow$} & 
\multicolumn{1}{>{\columncolor{lower_metric}}l}{Sq Rel$\downarrow$} &
\multicolumn{1}{>{\columncolor{lower_metric}}l}{RMSE$\downarrow$} & 
\multicolumn{1}{>{\columncolor{lower_metric}}l}{RMSE $\log\downarrow$} & \multicolumn{1}{>{\columncolor{higher_metric}}l}{$\delta < 1.25\uparrow$} &
\multicolumn{1}{>{\columncolor{higher_metric}}l}{$\delta < 1.25^2\uparrow$} &
\multicolumn{1}{>{\columncolor{higher_metric}}l}{$\delta < 1.25^3\uparrow$} \\
\hline 
Godard \emph{et al.} \cite{godard2019digging} & \cmark & S & 192$\times$640 & 0.108 & 0.842 & 4.891 & 0.207 & 0.866 & 0.949 & 0.976 \\
Chen \emph{et al.} \cite{chen2019towards} & \cmark & S,C(CS) & 256$\times$512 & 0.118 & 0.905 & 5.096 & 0.211 & 0.839 & 0.945 & 0.977 \\
Zhu \emph{et al.} \cite{zhu2020edge} & \cmark & S,C\textsuperscript{*} & 192$\times$640 & 0.102 & 0.804 & 4.629 & 0.189 & \underline{0.883} & \underline{0.961} & 0.981 \\
Bhat \emph{et al.} \cite{bhat2021adabins} & \cmark & S & 192$\times$640 & 0.108 & 0.909 & 4.874 & 0.199 & 0.870 & 0.953 & 0.978 \\
Chen \emph{et al.} \cite{chen2021aggnet} & \cmark & S & 192$\times$640 & 0.100 & 0.757 & 4.581 & 0.189 & 0.882 & \underline{0.961} & 0.981 \\
Peng \emph{et al.} \cite{peng2021excavating} & \cmark & S & 192$\times$640 & \underline{0.099} & 0.754 & 4.490 & 0.183 & 0.888 & {\bf0.963} & 0.982 \\
Poggi \emph{et al.} \cite{poggi2022real} & \xmark & S & 192$\times$640 & 0.127 & 1.059 & 5.259 & 0.218 & 0.834 & 0.942 & 0.974 \\
Li \emph{et al.} \cite{li2022self} & \cmark & S & 192$\times$640 & 0.100 & \underline{0.735} & 4.453 & {\bf0.178} & 0.881 & {\bf0.963} & {\bf0.984} \\
Cha \emph{et al.} \cite{cha2022self} & \xmark & S & 192$\times$640 & 0.106 & 0.817 & 4.838 & 0.199 & 0.871 & 0.953 & 0.978 \\
\cdashline{1-11}
\rowcolor{LightYellow}
Watson \emph{et al.} \cite{watson2019self} & \cmark & S & 192$\times$640 & 0.102 & 0.762 & 4.602 & 0.189 & 0.880 & 0.960 & 0.981 \\
\rowcolor{StandardYellow}
{\bf SemHint-MD(Ours)} & \cmark & S,C\textsuperscript{*} & 192$\times$640 & {\bf0.098} & {\bf0.675} & {\bf4.345} & \underline{0.180} & {\bf0.888} & {\bf0.963} & \underline{0.983} \\
\hline
Pillai \emph{et al.} \cite{pillai2019superdepth} & \xmark & S & 384$\times$1024 & 0.112 & 0.875 & 4.958 & 0.207 & 0.852 & 0.947 & 0.977 \\
Godard \emph{et al.} \cite{godard2019digging} & \cmark & S & 320$\times$1024 & 0.105 & 0.822 & 4.692 & 0.199 & 0.876 & 0.954 & 0.977 \\
Zhu \emph{et al.} \cite{zhu2020edge} & \cmark & S,C\textsuperscript{*} & 320$\times$1024 & {\bf0.091} & \underline{0.646} & 4.244 & 0.177 & 0.898 & {\bf0.966} & 0.983 \\ 
Chen \emph{et al.} \cite{chen2021aggnet} & \cmark & S & 320$\times$1024 & 0.095 & 0.710 & 4.392 & 0.184 & 0.892 & \underline{0.963} & 0.982 \\
Peng \emph{et al.} \cite{peng2021excavating} & \cmark & S & 320$\times$1024 & {\bf0.091} & \underline{0.646} & \underline{4.207} & \underline{0.176} & {\bf0.901} & {\bf0.966} & 0.983 \\
Li \emph{et al.} \cite{li2022self} & \cmark & S & 320$\times$1024 & \underline{0.093} & 0.652 & 4.256 & {\bf0.173} & 0.891 & {\bf0.966} & {\bf0.985} \\
\cdashline{1-11}
\rowcolor{LightYellow}
Watson \emph{et al.} \cite{watson2019self} & \cmark & S & 320$\times$1024 & 0.096 & 0.710 & 4.393 & 0.185 & 0.890 & 0.962 & 0.981 \\
\rowcolor{StandardYellow}
{\bf SemHint-MD(Ours)} & \cmark & S,C\textsuperscript{*} & 320$\times$1024 & {\bf0.091} & {\bf0.624} & {\bf4.181} & \underline{0.176} & \underline{0.899} & {\bf0.966} & \underline{0.984} \\
\hline
\multicolumn{11}{l}{\textit{PP} --- post-processing \cite{godard2017unsupervised}; in \emph{Data} column: \textit{S} --- self-supervised stereo supervision, \textit{C} --- ground truth semantic labels are used for training, \textit{C\textsuperscript{*}} --- pre}\\ 
\multicolumn{11}{l}{-trained model-predicted pseudo semantic labels are used for training, \textit{(CS)} --- the segmentation branch is trained on Cityscape \cite{cordts2016cityscapes}; each light yellow}\\ 
\multicolumn{11}{l}{-highlighted \colorbox{LightYellow}{row} is a baseline and the yellow-highlighted \colorbox{StandardYellow}{row} following it is our method that is developed from this baseline. Best results are in \textbf{bold},}\\
\multicolumn{11}{l}{second best are \underline{underlined}.}
\end{tabular}
\end{adjustbox}
\vspace{-1em}
\label{tab:kitti-rst}
\end{table*}

\begin{table*}[t!]
\caption{Performance comparisons of different baselines with and without our methods on KITTI Stereo 2015 dataset.}
\vspace{-1em}
\centering
\begin{adjustbox}{width=0.8\textwidth}
\begin{tabular}{l|cccccccccc}
\hline
Method & \multicolumn{1}{>{\columncolor{lower_metric}}l}{Abs Rel$\downarrow$} & 
\multicolumn{1}{>{\columncolor{lower_metric}}l}{Sq Rel$\downarrow$} &
\multicolumn{1}{>{\columncolor{lower_metric}}l}{RMSE$\downarrow$} & 
\multicolumn{1}{>{\columncolor{lower_metric}}l}{RMSE $\log\downarrow$} & \multicolumn{1}{>{\columncolor{higher_metric}}l}{$\delta < 1.25\uparrow$} &
\multicolumn{1}{>{\columncolor{higher_metric}}l}{$\delta < 1.25^2\uparrow$} &
\multicolumn{1}{>{\columncolor{higher_metric}}l}{$\delta < 1.25^3\uparrow$} \\
\hline
\rowcolor{LightYellow}
Watson \emph{et al.} \cite{watson2019self} (ResNet18) & 0.107 & 0.808 & 4.690 & 0.191 & 0.874 & 0.958 & 0.981 \\
\rowcolor{StandardYellow}
{\bf SemHint-MD(Ours)} & {\bf0.103} & {\bf0.717} & {\bf4.522} & {\bf0.185} & {\bf0.879} & {\bf0.961} & {\bf0.983} \\
\hline
\rowcolor{LightYellow}
Watson \emph{et al.} \cite{watson2019self} (ResNet50) & 0.102 & 0.762 & 4.602 & 0.189 & 0.880 & 0.960 & 0.981 \\
\rowcolor{StandardYellow}
{\bf SemHint-MD(Ours)} & {\bf0.098} & {\bf0.675} & {\bf4.345} & {\bf0.180} & {\bf0.888} & {\bf0.963} & {\bf0.983}\\
\hline
\rowcolor{LightYellow}
Bhat \emph{et al.} \cite{bhat2021adabins} (ResNet18) & {\bf0.108} & 0.864 & 4.799 & 0.194 & {\bf0.872} & 0.957 & 0.980 \\
\rowcolor{StandardYellow}
{\bf SemHint-MD(Ours)} & {\bf0.108} & {\bf0.784} & {\bf4.624} & {\bf0.187} & 0.870 & {\bf0.960} & {\bf0.983}\\
\hline
\rowcolor{LightYellow}
Bhat \emph{et al.} \cite{bhat2021adabins} (ResNet50) & 0.108 & 0.909 & 4.874 & 0.199 & 0.870 & 0.953 & 0.978 \\
\rowcolor{StandardYellow}
{\bf SemHint-MD(Ours)} & \textbf{0.104} & \textbf{0.762} & \textbf{4.557} & \textbf{0.184} & \textbf{0.878} & \textbf{0.961} & \textbf{0.983} \\
\hline
\multicolumn{8}{l}{All models are run at resolution 192$\times$640 and post-processing \cite{godard2017unsupervised} is used. See Table \ref{tab:kitti-rst} for other table contents.}
\end{tabular}
\end{adjustbox}
\label{tab:kitti-baseline-rst}
\vspace{-1em}
\end{table*}

\section{Experiments} \label{sec:exp}

We demonstrate the performance of the proposed framework on the KITTI benchmark. 
We further show that SemHint-MD can benefit downstream tasks by deploying it to a tissue deformation tracking and reconstruction task on a robotic surgery platform. 
Our models are implemented in PyTorch \cite{paszke2019pytorch} and trained on one NVIDIA GeForce RTX 3090 GPU. 



\subsection{Evaluation on KITTI dataset}
\subsubsection{Baseline and Implementation Details} We use KITTI Stereo 2015 dataset \cite{geiger2012we} under the data split of Eigen \emph{et al.} \cite{eigen2014depth}. 
Following \cite{zhu2020edge}, the pseudo-semantic ground truth of all training data is generated offline by a network that was trained on only 200 domain images \cite{zhu2019improving}. 
We take \cite{watson2019self} as our baseline, and
$L_{rfd}$ and $L_rfs$ are only used when the models can provide good enough depth and segmentation. 
Specifically, the models are first trained for 20 epochs on stereo image pairs using the Adam optimizer \cite{kingma2014adam}, with a batch size of 12, and an initial learning rate of $10^{-4}$, which is step decayed by 0.1 every 5 epochs. 
During this process, we set $\alpha=0.5$, $c_1=0.1$, $c_2=100$, $\beta_1=1$, $\beta_2=1$, $\lambda_{pe}=1$, $\lambda_h=1$, $\lambda_{rfd}=0$, $\lambda_s=0$, $\lambda_{ps}=1$, and $\lambda_{rfs}=0$. 
After the models converge at the 20th epoch, we continue to train them for 5 epochs, with $\lambda_{rfd}=1$, $\lambda_{rfs}=1$, and a learning rate of $10^{-6}$ for the first epoch and $10^{-7}$ for the remainder, while other parameters remain the same. 
For testing, the predicted depth maps are evaluated with standard seven KITTI metrics and a standard depth cap of 80 meters \cite{godard2017unsupervised}. 

\begin{figure*}[!t]
\centering
\includegraphics[width=0.98\textwidth]{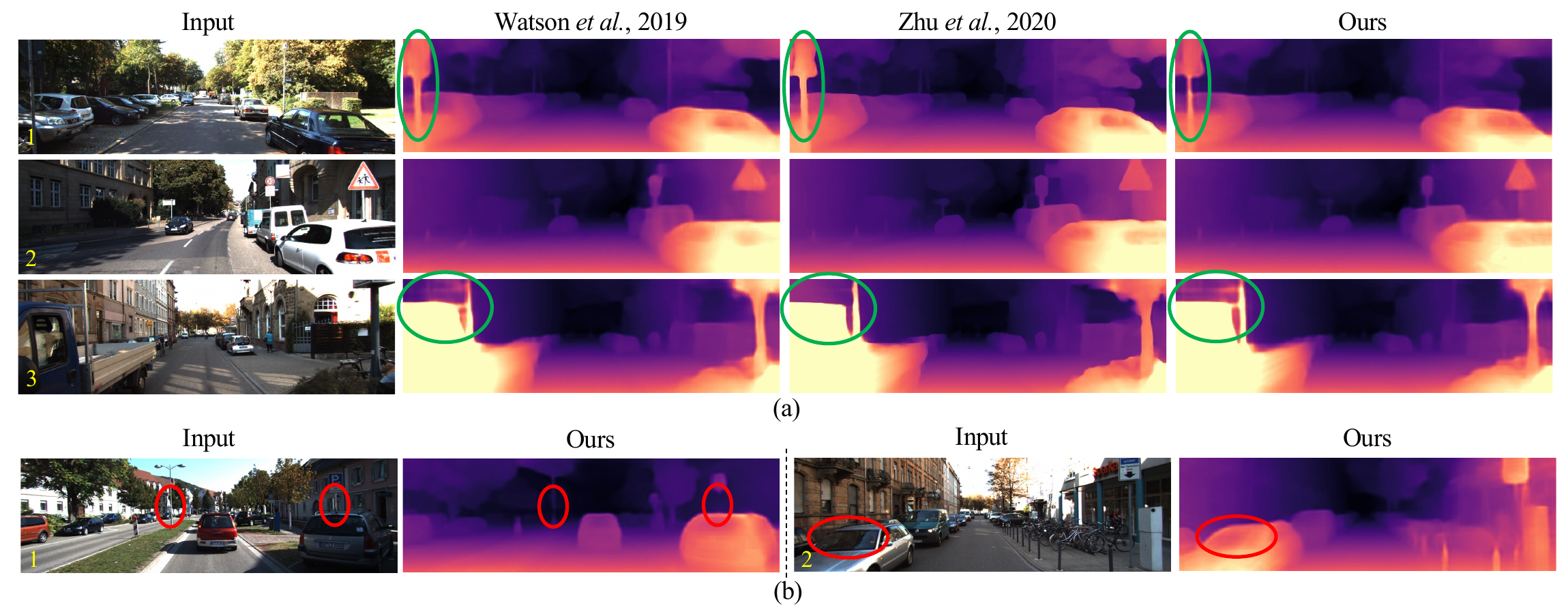}
\vspace{-1em}
\caption{Qualitative results on KITTI Stereo 2015 Eigen split. (a) Input video frames and predicted depth maps. (b) Failure cases. Additional results can be found in Appendix \ref{appx:kitti_exp}.}
\vspace{-1.5em}
\label{fig:kitti_sample}
\end{figure*}


\subsubsection{Quantitative Analysis}
Results at low resolution 
and high resolution 
are shown in Table \ref{tab:kitti}. SemHint-MD achieves SOTA performance at both low and high resolutions. 
It should be note that the predicted depth map is refined only in the less reliable regions, such as semantic boundaries, so the number of pixels influenced by mutual refining is relatively small. Thus, \emph{$\delta < 1.25$}, \emph{$\delta < 1.25^2$}, and \emph{$\delta < 1.25^3$} remain close as they measure the percentage of pixels that have depth lies within a certain ratio to the ground truth, while we see a clear decrease in \emph{Abs Rel}, \emph{Sq Rel}, \emph{RMSE}, and \emph{RMSE log}, meaning SemHint-MD achieves considerable improvements in regions that were difficult to estimate the depth. 
Also, the proposed methods can be flexibly integrated with general depth estimation or MTL models. 
Table \ref{tab:kitti-baseline-rst} shows the performance of SemHint-MD when integrated with different baselines, including a transformer \cite{bhat2021adabins}, and the proposed methods improve all involved baselines. 

\subsubsection{Qualitative Analysis}
Fig. \ref{fig:kitti_sample} shows examples of predicted depth, where Zhu \emph{et al.} \cite{zhu2020edge} also used Watson \emph{et al.} \cite{watson2019self} as the baseline.  
As explained in Section \ref{sec:related_work}, ``bleeding artifact" leads to a foreground object region in the depth map larger than its true shape, while a better depth should be well aligned with the object. 
Our method generally produces depth maps that better match object regions, the slim traffic sign pole in Fig. \ref{fig:kitti_sample}(a).1 is an example. 
In addition, when the network is trapped by a local minimum, it may fail to find the correspondences of some small or slim objects between different views. 
The horizontal truck rack in Fig. \ref{fig:kitti_sample}(a).3 is a typical example of such challenging objects and its depth given by the two comparative methods is almost ``fused" into the background, 
while our method can identify this region and use Algorithm \ref{alg:seg_help_depth} to produce depth hints for it, so we can see our method provides better depth for the truck rack. 

Fig. \ref{fig:kitti_sample}(b) shows typical failure cases of SemHint-MD. It can still fail on small objects, especially for those that usually get wrong semantic predictions. Fig. \ref{fig:kitti_sample}(b).1 shows such a situation where the faraway poles are ``fused" into the background. 
As shown in Fig. \ref{fig:kitti_sample}(b).2, current methods, including our baselines, usually give car glass a larger depth than it is supposed to be. For some boundary points of the car that are considered less reliable, our method may find depth from the glass as depth hints, which leads to a larger erroneous depth region. 

\begin{table*}[t]
\caption{Ablation study on KITTI Stereo 2015 dataset.}
\vspace{-1em}
\centering
\begin{adjustbox}{width=\textwidth}
\begin{tabular}{c|ccc|cc|cccccccc}
\hline
ID & Share Dec. & Refine Depth & Refine Seg. & MACs(G) & Params(M) & \multicolumn{1}{>{\columncolor{lower_metric}}l}{Abs Rel$\downarrow$} & 
\multicolumn{1}{>{\columncolor{lower_metric}}l}{Sq Rel$\downarrow$} &
\multicolumn{1}{>{\columncolor{lower_metric}}l}{RMSE$\downarrow$} & 
\multicolumn{1}{>{\columncolor{lower_metric}}l}{RMSE $\log\downarrow$} & \multicolumn{1}{>{\columncolor{higher_metric}}l}{$\delta < 1.25\uparrow$} &
\multicolumn{1}{>{\columncolor{higher_metric}}l}{$\delta < 1.25^2\uparrow$} &
\multicolumn{1}{>{\columncolor{higher_metric}}l}{$\delta < 1.25^3\uparrow$} \\
\hline
1 & \xmark & \xmark & \xmark & 16.598 & 32.520 & 0.103 & 0.790 & 4.611 & 0.188 & 0.879 & 0.960 & 0.982 \\
\hline
2 & \cmark & \xmark & \xmark & \multirow{4}{*}{38.772} & \multirow{4}{*}{33.242} & 0.100 & 0.752 & 4.492 & 0.181 & 0.886 & 0.963 & 0.983 \\
3 & \cmark & \cmark$^*$ & \xmark & & & 0.100 & 0.710 & 4.389 & {\bf0.180} & 0.887 & {\bf0.964} & {\bf0.984}\\
4 & \cmark & \cmark & \xmark & & & 0.100 & 0.708 & 4.388 & {\bf0.180} & 0.887 & 0.963 & {\bf0.984} \\
5 & \cmark & \cmark & \cmark  & & & {\bf0.098} & {\bf0.675} & {\bf4.345} & {\bf0.180} & {\bf0.888} & 0.963 & 0.983 \\
\hline\hline
6 & $l_0$ & \multirow{5}{*}{\xmark} & \multirow{5}{*}{\xmark} & 45.003 & 42.252 & {\bf0.100} & 0.754 & 4.540 & 0.184 & 0.885 & 0.962 & {\bf0.983} \\
7 & $l_1$ & & & 42.880 & 34.288 & 0.101 & 0.762 & 4.534 & 0.184 & 0.884 & 0.962 & 0.982 \\
8 & $l_2$ & & & 41.322 & 33.477 & 0.101 & 0.769 & 4.549 & 0.184 & {\bf0.886} & 0.962 & {\bf0.983} \\
9 & $l_3$ & & & 39.764 & 33.274 & 0.101 & 0.765 & 4.541 & 0.182 & 0.884 & {\bf0.963} & {\bf0.983}\\
10 & $l_4$ & & & 38.772 & 33.242 & \textbf{0.100} & \textbf{0.752} & \textbf{4.492} & \textbf{0.181} & \textbf{0.886} & \textbf{0.963} & \textbf{0.983} \\
\hline
\multicolumn{14}{l}{Baseline: Watson \emph{et al.} \cite{watson2019self} with ResNet50 as the encoder, runing at resolution 192$\times$640. \emph{Share Dec.} --- share decoder; \emph{Refine Depth} --- refine depth with segmentation, where}\\
\multicolumn{14}{l}{\cmark$*$ --- pseudo-semantic ground truth is used to refine depth; \emph{Refine Seg.} --- refine segmentation with depth; \emph{MACs} --- number of multiply-accumulate operations \cite{whitehead2011precision}, it reveals}\\
\multicolumn{14}{l}{the computational burden of the models; \emph{Params} --- number of model parameters, it reveals the model size. See Table \ref{tab:kitti-rst} for other table contents.}
\end{tabular}
\end{adjustbox}
\label{tab:kitti_ablation}
\end{table*}

\subsubsection{Ablation Study}

We conduct ablation studies to understand how 1) sharing decoder layers, 2) refining depth with segmentation, and 3) refining segmentation with depth contribute to the overall performance, as shown in Table \ref{tab:kitti_ablation}. 
When \emph{Refine Depth} is selected, $L_{rfd}$ is used to supervise training. 
Compare Row 2 with Row 3 and 4, we find using either the pseudo-semantic ground truth or the segmentation predictions to refine depth can enhance the performance. 
Comparing Row 4 with Row 5, we find using the depth to refine segmentation, which is then used to refine depth, can further improve the performance. 
Row 6-10 show the model performance when different numbers of decoder parameters are shared. 
From $l_0$ to $l_4$, the number of shared decoder parameters is increasing 
(see Appendix \ref{appx:network_mono2} for structures of $l_0$-$l_4$). 
The result shows the depth and segmentation branches have many common features that allow deeper model sharing, which leads to smaller computation cost and model size. 

\subsection{Downstream Endoscopic Tissue Tracking Task} \label{sec:super}

We further integrate SemHint-MD with a {\bf Su}rgical {\bf Per}ception framework (SuPer) \cite{li2020super} that can densely track points on the tissue surface in the 3D space. 

\subsubsection{Data and Implementation Details} SuPer is evaluated on its own data, an endoscopic video collected using the da Vinci Research Kit (dVRK) \cite{kazanzides2014open, richter2021bench} from an \emph{ex vivo} scene, where a piece of chicken meat was manipulated by the griper of dVRK. The video consists of 520 stereo frames, which were rectified to a resolution of 640$\times$480. 
To quantitatively evaluate the tracking performance, the trajectories of 20 points on the tissue surface were labeled throughout the video. We project the tracked points to the image plane and calculate their reprojection errors as the distances between the reprojections and the corresponding ground truth. 

The learning models for depth estimation and segmentation are trained on other 320 video frames collected from the same scene. 
Similar to the learning routine of KITTI data, 
the models are first trained for 60 epochs with $\alpha=0.5$, $c_1=0.1$, $c_2=80$, $\beta_1=1$, $\beta_2=1$, $\lambda_{pe}=1$, $\lambda_h=0$, $\lambda_{rfd}=0$, $\lambda_s=0.001$, $\lambda_{ps}=1$, $\lambda_{rfs}=0$, a batch size of 6, and an initial learning rate of $10^{-4}$, which is step decayed by 0.1 every 15 epochs.  
Then we continue to train the models for 5 epochs, with $\lambda_{rfd}=1$, $\lambda_{rfs}=1$, and a learning rate of $10^{-6}$ for the first epoch and $10^{-7}$ for the remainder, while other parameters remain the same. 

\subsubsection{Results} SuPer takes depth maps of the endoscopic video frames as inputs and uses semantic maps of the scene to extract points on the tissue surface to track. We compare the performance of SuPer under two different ways of obtaining its inputs: 1) {\bf Separate}: the depth and segmentation maps are estimated by separately trained depth branch and segmentation branch of SemHint-MD, respectively; and 2) {\bf SemHint-MD}: the depth and segmentation maps are estimated at the same time by SemHint-MD. The results are shown in Table \ref{tab:super}, SemHint-MD can generate the inputs for SuPer with less time and lead to more accurate tissue tracking.

\begin{table}[t]
\caption{Performance on SuPer data.} 
\vspace{-1em}
\centering
\begin{adjustbox}{width=0.9\linewidth}
\begin{tabular}{lcc}
\hline
Input & \multicolumn{1}{>{\columncolor{lower_metric}}l}{Inf. Time (ms)$\downarrow$} & \multicolumn{1}{>{\columncolor{higher_metric}}l}{Reprojection Error (pixel)$\uparrow$} \\
\hline
Separate & 24.1 & 12.4$\pm$12.0\\
\hline
SemHint-MD & {\bf11.9} & {\bf10.0$\pm$9.0}\\
\hline
\multicolumn{3}{l}{\emph{Inf. Time} denotes the inference time taken by the learning models}\\
\multicolumn{3}{l}{to generate inputs, \emph{i.e.}, the depth and segmentation maps, from}\\
\multicolumn{3}{l}{one video frame for SuPer. 
The reprojection errors are shown in}\\
\multicolumn{3}{l}{`mean$\pm$standard deviation'.}
\end{tabular}
\end{adjustbox} 
\label{tab:super}
\vspace{-2em}
\end{table}

\section{Conclusion}

Targeting the limitations of current self-supervised losses for depth estimation, we build an MTL model SemHint-MD and propose novel methods to let the current depth and semantic predictions help refine each other and use the refined depth and segmentation maps to provide extra supervision. 
Our extensive experiments on the KITTI and surgical data demonstrate that SemHint-MD can achieve SOTA performance. 
In the future, the proposed method can be integrated with domain adaptation methods to achieve better performance with limited data.


\bibliographystyle{IEEEtran}
\bibliography{ralbib}

\clearpage
\newpage
\appendix
\section{Appendix}
\subsection{MTL Network Architecture} \label{appx:network}

\subsubsection{MTL Networks Developed based on Monodepth2} \label{appx:network_mono2}
Our models all use either ResNet18 or ResNet50 \cite{he2016deep} as the encoder. The decoder architecture of the depth branch is the same as \cite{godard2019digging}, as shown in Table \ref{appxtab:mono2_dec}. 
The decoder of the segmentation branch shares some decoder layers with the depth branch. We define 5 levels of decoder parameter sharing: $l_0$ (only share encoder; shown in Table \ref{appxtab:segbranchl0}), $l_1$ (shown in Table \ref{appxtab:segbranchl1}), $l_2$ (shown in Table \ref{appxtab:segbranchl2}), $l_3$ (shown in Table \ref{appxtab:segbranchl3}), and $l_4$ (shown in Table \ref{appxtab:segbranchl4}).

\begin{table}[htbp!]
\caption{Depth branch decoder architecture \cite{godard2019digging}.}
\begin{adjustbox}{width=0.49\textwidth}
\begin{tabular}{|c|c|c|c|c|c|}
\hline
Layer Name & Input  & k & s & chn & Activation \\
\hline
upconv5    & econv5 & 3 & 1 & 256 & ELU \\
iconv5 & $\uparrow$upconv5, econv4 & 3 & 1 & 256 & ELU \\
\hline
upconv4    & iconv5 & 3 & 1 & 128 & ELU \\
iconv4 & $\uparrow$upconv4, econv3 & 3 & 1 & 128 & ELU \\
disp4 & iconv4 & 3 & 1 & 1 & Sigmoid \\
\hline
upconv3    & iconv4 & 3 & 1 & 64 & ELU \\
iconv3 & $\uparrow$upconv3, econv2 & 3 & 1 & 64 & ELU \\
disp3 & iconv3 & 3 & 1 & 1 & Sigmoid \\
\hline
upconv2    & iconv3 & 3 & 1 & 32 & ELU \\
iconv2 & $\uparrow$upconv2, econv1 & 3 & 1 & 32 & ELU \\
disp2 & iconv2 & 3 & 1 & 1 & Sigmoid \\
\hline
upconv1    & iconv2 & 3 & 1 & 16 & ELU \\
iconv1 & $\uparrow$upconv1 & 3 & 1 & 16 & ELU \\
disp1 & iconv1 & 3 & 1 & 1 & Sigmoid \\
\hline
\multicolumn{6}{l}{\emph{k} --- kernel size, \emph{s} --- stride, \emph{chn} --- output channel number, \emph{econv} --- encoder}\\
\multicolumn{6}{l}{features, \emph{disp} --- output disparity map, $\uparrow$ --- a 2$\times$ nearest-neighbor upsample}\\
\multicolumn{6}{l}{operation.}
\end{tabular}
\end{adjustbox}
\label{appxtab:mono2_dec}
\end{table}

\begin{table*}[htbp!]
\caption{Semantic segmentation branch architecture under $l_0$ configuration.} 
\centering
\begin{adjustbox}{width=0.99\textwidth}
\begin{tabular}{|c|c|c|c|c|c|c|}
\hline
Layer Name & Input  & k & s & chn & BatchNorm & Activation \\
\hline
upsconv5    & econv5 & 3 & 1 & 256 & \xmark & ELU \\
isconv5 & $\uparrow$upsconv5, econv4 & 3 & 1 & 256 & \xmark & ELU \\
\hline
upsconv4    & isconv5 & 3 & 1 & 128 & \xmark& ELU \\
isconv4 & $\uparrow$upsconv4, econv3 & 3 & 1 & 128 & \xmark& ELU \\
\hline
upsconv3    & isconv4 & 3 & 1 & 64 & \xmark& ELU \\
isconv3 & $\uparrow$upsconv3, econv2 & 3 & 1 & 64 & \xmark& ELU \\
\hline
upsconv2    & isconv3 & 3 & 1 & 32 & \xmark& ELU \\
isconv2 & $\uparrow$upsconv2, econv1 & 3 & 1 & 32 & \xmark& ELU \\
\hline
upsconv1    & isconv2 & 3 & 1 & 16 & \xmark& ELU \\
isconv1 & $\uparrow$upsconv1 & 3 & 1 & 16 & \xmark & ELU \\
\hline
sconv1    & $\uparrow^{\times 16}$upsconv5, $\uparrow^{\times 8}$upsconv4, $\uparrow^{\times 4}$upsconv3, $\uparrow^{\times 2}$upsconv2, upsconv1 & 3 & 1 & 128 & \cmark & ReLU\\
sconv2 & sconv1 & 3 & 1 & 128 & \cmark & ReLU\\
sconv3 & sconv2 & 1 & 1 & 1 & \xmark & Softmax\\
\hline
\multicolumn{7}{l}{\emph{upsconv}, \emph{isconv}, and \emph{sconv} --- segmentation branch-specific convolution layers, and $\uparrow^{\times n}$ --- a n$\times$ bilinear upsampling operation. Refer to Table \ref{appxtab:mono2_dec} for}\\
\multicolumn{7}{l}{other table contents.}
\end{tabular}
\end{adjustbox}
\label{appxtab:segbranchl0}
\end{table*}

\begin{table*}[htbp!]
\caption{Semantic segmentation branch architecture under $l_1$ configuration.}
\centering
\begin{adjustbox}{width=0.99\textwidth}
\begin{tabular}{|c|c|c|c|c|c|c|}
\hline
Layer Name & Input  & k & s & chn & BatchNorm & Activation \\
\hline
upsconv4    & iconv5 & 3 & 1 & 128 & \xmark& ELU \\
isconv4 & $\uparrow$upsconv4, econv3 & 3 & 1 & 128 & \xmark& ELU \\
\hline
upsconv3    & isconv4 & 3 & 1 & 64 & \xmark& ELU \\
isconv3 & $\uparrow$upsconv3, econv2 & 3 & 1 & 64 & \xmark& ELU \\
\hline
upsconv2    & isconv3 & 3 & 1 & 32 & \xmark& ELU \\
isconv2 & $\uparrow$upsconv2, econv1 & 3 & 1 & 32 & \xmark& ELU \\
\hline
upsconv1    & isconv2 & 3 & 1 & 16 & \xmark& ELU \\
isconv1 & $\uparrow$upsconv1 & 3 & 1 & 16 & \xmark & ELU \\
\hline
sconv1    & $\uparrow^{\times 16}$upconv5, $\uparrow^{\times 8}$upsconv4, $\uparrow^{\times 4}$upsconv3, $\uparrow^{\times 2}$upsconv2, upsconv1 & 3 & 1 & 128 & \cmark & ReLU\\
sconv2 & sconv1 & 3 & 1 & 128 & \cmark & ReLU\\
sconv3 & sconv2 & 1 & 1 & 1 & \xmark & Softmax\\
\hline
\multicolumn{7}{l}{Refer to Table \ref{appxtab:mono2_dec} and Table \ref{appxtab:segbranchl0} for table contents.}
\end{tabular}
\end{adjustbox}
\label{appxtab:segbranchl1}
\end{table*}

\begin{table*}[htbp!]
\caption{Semantic segmentation branch architecture under $l_2$ configuration.}
\centering
\begin{adjustbox}{width=0.99\textwidth}
\begin{tabular}{|c|c|c|c|c|c|c|}
\hline
Layer Name & Input  & k & s & chn & BatchNorm & Activation \\
\hline
upsconv3    & iconv4 & 3 & 1 & 64 & \xmark& ELU \\
isconv3 & $\uparrow$upsconv3, econv2 & 3 & 1 & 64 & \xmark& ELU \\
\hline
upsconv2    & isconv3 & 3 & 1 & 32 & \xmark& ELU \\
isconv2 & $\uparrow$upsconv2, econv1 & 3 & 1 & 32 & \xmark& ELU \\
\hline
upsconv1    & isconv2 & 3 & 1 & 16 & \xmark& ELU \\
isconv1 & $\uparrow$upsconv1 & 3 & 1 & 16 & \xmark & ELU \\
\hline
sconv1    & $\uparrow^{\times 16}$upconv5, $\uparrow^{\times 8}$upconv4, $\uparrow^{\times 4}$upsconv3, $\uparrow^{\times 2}$upsconv2, upsconv1 & 3 & 1 & 128 & \cmark & ReLU\\
sconv2 & sconv1 & 3 & 1 & 128 & \cmark & ReLU\\
sconv3 & sconv2 & 1 & 1 & 1 & \xmark & Softmax\\
\hline
\multicolumn{7}{l}{Refer to Table \ref{appxtab:mono2_dec} and Table \ref{appxtab:segbranchl0} for table contents.}
\end{tabular}
\end{adjustbox}
\label{appxtab:segbranchl2}
\end{table*}

\begin{table*}[htbp!]
\caption{Semantic segmentation branch architecture under $l_3$ configuration.}
\centering
\begin{adjustbox}{width=0.99\textwidth}
\begin{tabular}{|c|c|c|c|c|c|c|}
\hline
Layer Name & Input  & k & s & chn & BatchNorm & Activation \\
\hline
upsconv2    & iconv3 & 3 & 1 & 32 & \xmark& ELU \\
isconv2 & $\uparrow$upsconv2, econv1 & 3 & 1 & 32 & \xmark& ELU \\
\hline
upsconv1    & isconv2 & 3 & 1 & 16 & \xmark& ELU \\
isconv1 & $\uparrow$upsconv1 & 3 & 1 & 16 & \xmark & ELU \\
\hline
sconv1    & $\uparrow^{\times 16}$upconv5, $\uparrow^{\times 8}$upconv4, $\uparrow^{\times 4}$upconv3, $\uparrow^{\times 2}$upsconv2, upsconv1 & 3 & 1 & 128 & \cmark & ReLU\\
sconv2 & sconv1 & 3 & 1 & 128 & \cmark & ReLU\\
sconv3 & sconv2 & 1 & 1 & 1 & \xmark & Softmax\\
\hline
\multicolumn{7}{l}{Refer to Table \ref{appxtab:mono2_dec} and Table \ref{appxtab:segbranchl0} for table contents.}
\end{tabular}
\end{adjustbox}
\label{appxtab:segbranchl3}
\end{table*}

\begin{table*}[htbp!]
\caption{Semantic segmentation branch architecture under $l_4$ configuration.}
\centering
\begin{adjustbox}{width=0.99\textwidth}
\begin{tabular}{|c|c|c|c|c|c|c|}
\hline
Layer Name & Input  & k & s & chn & BatchNorm & Activation \\
\hline
sconv1    & $\uparrow^{\times 16}$upconv5, $\uparrow^{\times 4}$upconv4, $\uparrow^{\times 8}$upconv3, $\uparrow^{\times 2}$upconv2, upconv1 & 3 & 1 & 128 & \cmark & ReLU \\
sconv2 & sconv1 & 3 & 1 & 128 & \cmark & ReLU\\
sconv3 & sconv2 & 1 & 1 & 1 & \xmark & Softmax\\
\hline
\multicolumn{7}{l}{Refer to Table \ref{appxtab:mono2_dec} and Table \ref{appxtab:segbranchl0} for table contents.}
\end{tabular}
\end{adjustbox}
\label{appxtab:segbranchl4}
\end{table*}

\subsection{More Experiments} \label{appx:exp}
\subsubsection{Evaluation on KITTI Dataset}\label{appx:kitti_exp}
\begin{figure}[htbp!]
\centering
\includegraphics[width=\linewidth]{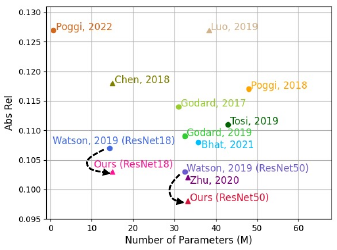}
\vspace{-2.em}
\caption{Visualization of model size and depth estimation performance on KITTI Stereo 2015. Methods marked by circles ($\bullet$) are only for depth estimation, and methods marked by triangles ($\blacktriangle$) predict both depth and segmentation maps. The black dashed arrow starts from the baseline \cite{watson2019self} and points to our method with this baseline. Our method achieves superior performance while not increasing the model size too much. }
\vspace{-2em}
\label{fig:kitti_params}
\end{figure}

\begin{figure*}[htbp!]
\centering
\vspace{-0.5em}
\includegraphics[width=0.99\textwidth]{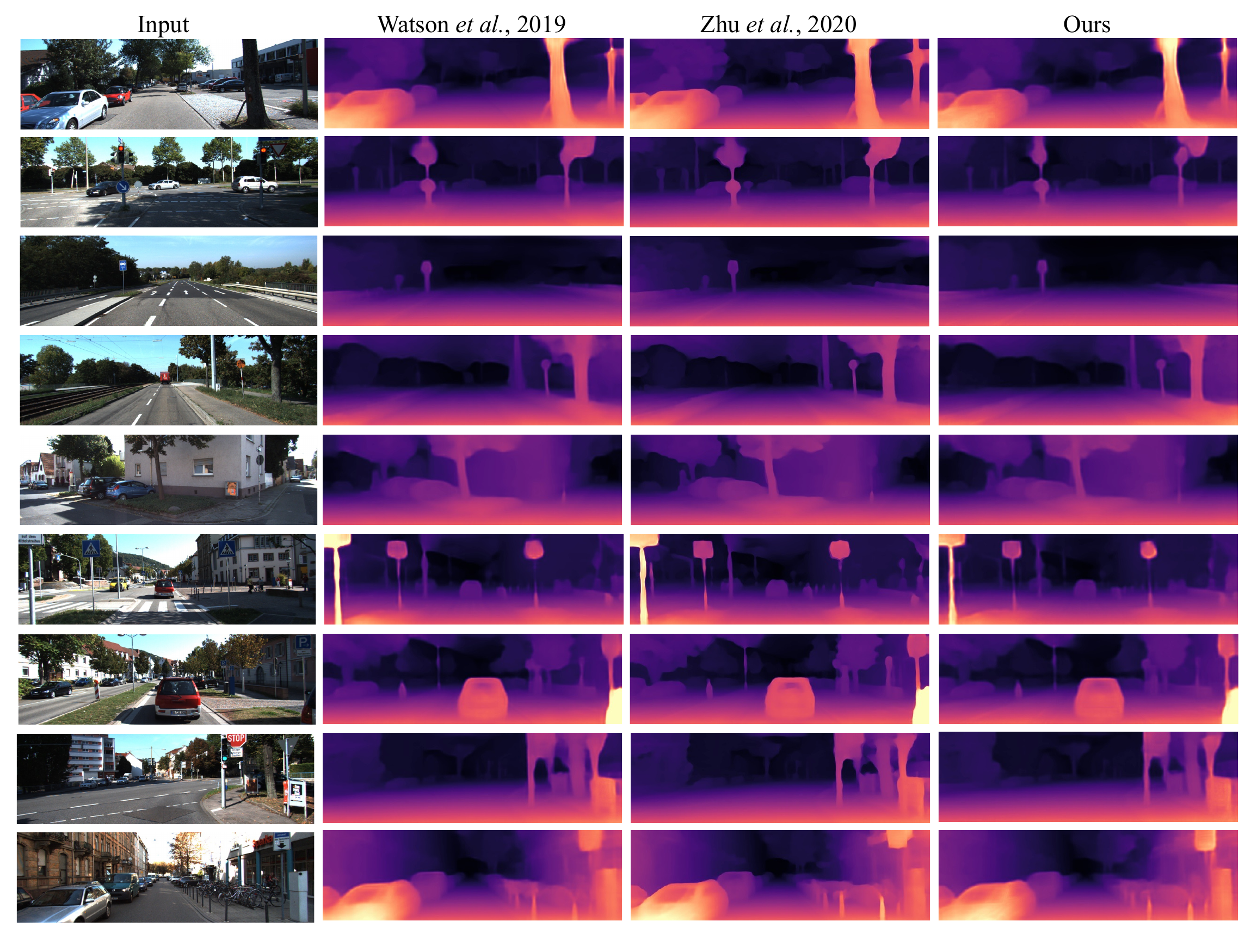}
\vspace{-1.em}
\caption{More qualitative results on KITTI Stereo 2015 Eigen split.}
\vspace{-1.5em}
\label{appxfig:kitti_more_sample}
\end{figure*}

We also compare the parameter number of our method and SOTA methods with corresponding depth estimation performance in Fig. \ref{fig:kitti_params}. 
Each black arrow in Fig. \ref{fig:kitti_params} starts from a baseline and points to our method developed from this baseline. Most methods are only for depth estimation (shown in circles). In contrast, our method achieves superior depth performance while only adding a very small number of parameters to build the segmentation branch. 
More qualitative results are shown in Fig. \ref{appxfig:kitti_more_sample}.

\end{document}